\documentclass[conference]{IEEEtran}
\IEEEoverridecommandlockouts
\usepackage{cite}
\usepackage{amsmath,amssymb,amsfonts}
\usepackage{algorithmic}
\usepackage{graphicx}
\usepackage{textcomp}
\usepackage{hyperref}       
\usepackage{url}            
\usepackage{booktabs}       
\usepackage{amsfonts}       
\usepackage{nicefrac}       
\usepackage{microtype}      
\usepackage{xcolor}         
\usepackage{pgfplots}  
\usepackage{graphicx}
\graphicspath{ {./} }
\def\BibTeX{{\rm B\kern-.05em{\sc i\kern-.025em b}\kern-.08em
    T\kern-.1667em\lower.7ex\hbox{E}\kern-.125emX}}
\begin{document}

\title{Exploring the Knowledge Mismatch Hypothesis: \\ Hallucination Propensity in Small Models Fine-tuned on Data from Larger Models\\

\author{\IEEEauthorblockN{ Phil Wee}
\IEEEauthorblockA{\textit{Computer Science} \\
\textit{NYU}\\
Abu Dhabi, UAE \\
philwee@nyu.edu}
\and
\IEEEauthorblockN{Riyadh Baghdadi}
\IEEEauthorblockA{\textit{Computer Science} \\
\textit{NYU}\\
Abu Dhabi, UAE \\
baghdadi@nyu.edu}
}
}
\maketitle

\begin{abstract}
Recently, there has been an explosion of large language models created through fine-tuning with data from larger models. These small models able to produce outputs that appear qualitatively similar to significantly larger models. However, one of the key limitations that have been observed with these models is their propensity to hallucinate significantly more often than larger models. In particular, they have been observed to generate coherent outputs that involve factually incorrect information and spread misinformation, toxicity, and stereotypes. There are many potential causes of hallucination, of which, one hypothesis is that fine-tuning a model on data produced by a larger model leads to a knowledge mismatch which contributes to hallucination. In particular, it is hypothesized that there is a mismatch between the knowledge that is fed to the model to fine-tune it and the knowledge that is already present in the graph. Fine-tuning the model on data that has such mismatch could contribute to an increased propensity to hallucinate. We show that on an unseen test set, a smaller model fine-tuned on data generated from a larger model produced more wrong answers when compared to models fine-tuned on data created by the small model, which confirms the hypothesis. 
\end{abstract}

\begin{IEEEkeywords}
large language models, fine-tuning, hallucination, evaluation
\end{IEEEkeywords}

\section{Introduction}

A well-known limitation of large language models is their ability to hallucinate or generate factually incorrect statements ~\cite{dziri2022origin}. While large language models are often able to appear fluent, the responses generated by these systems have been observed to often produce statements that are misleading, factually incorrect and harmful. In the context of dialogue-based systems, it has been observed that models often produce responses that are not supported by the evidence available to the system ~\cite{dziri2022evaluating}. In the context of generative question and answering systems, models have been observed to provide factually incorrect responses ~\cite{Ji_2023}. This is a key issue in the field of Large Language Models because of the potential harm that hallucination can pose to users. For instance, in medical applications, a medical summary generated from patient information could pose a risk to the patient if the generated outputs involve hallucination. In this case, a factually false recommendation generated by the model can lead to a life-threatening incident for the patient.

In recent times, this is even more concerning because of the rise of strong large language models that are often able to produce coherent and semantically convincing responses while at the same time, involving hallucination in their outputs. It has become increasingly difficult to tell the difference between the outputs of a model and text written by humans, especially in more general and generic topics and in prompts with short responses. This furthers the risk of hallucination in the outputs of models - the high level of coherence in the outputs of many recent models can easily fool many humans and even many experts.

A key development in the field of open source models is the explosion of large language models created through fine-tuning with data from larger models. \cite{alpaca} \cite{xu2023baize} ~\cite{gpt4all} ~\cite{koala_blogpost_2023} ~\cite{vicuna2023} Most of these fine-tuned models are not only relatively small and easy to reproduce but are also relatively performant and are able to produce outputs that appear qualitatively similar to significantly larger models. However, one key limitation that many of these models share is their propensity to hallucinate significantly more often than larger models. While there has been a lot of previous work in addressing hallucination through further tuning via techniques such as Reinforcement Learning through Human Feedback (RLHF) ~\cite{ouyang2022training} ~\cite{bai2022training} and Reinforcement Learning through AI Feedback (RLAI) ~\cite{bai2022constitutional}, there has been relatively little work in analyzing the role of underlying training techniques such as fine-tuning.

Hallucination has a lot of possible reasons, both with respect to data and the model. Our goal in this paper is to study one of the possible reasons: fine-tuning data. In particular, we want to verify the hypothesis that fine-tuning a model on data produced by a larger model leads to a knowledge mismatch, which may contribute to hallucination. Our analysis finds that on an unseen test set, a smaller model fine-tuned on data generated from a larger model produced more wrong answers when compared to models fine-tuned on data created by the small model, which confirms the hypothesis. Our findings question the overall effectiveness of current fine-tuning practices, given the potential for a knowledge mismatch that such fine-tuning may cause and its implications on model hallucination.

\section{Origins of Hallucination}
Hallucination is observed to have many potential origins and is observed to be rooted in both data and models ~\cite{dziri2022origin} ~\cite{Ji_2023}. With respect to data, factually inaccurate data is known to be a key driver of hallucination - if the answer that the model was trained on was wrong, it is likely going to produce a wrong answer. However, even if the training data is factually accurate, hallucination can also still occur as there is simply no way to fully cover every single possible question in the training data. As a result, for questions that it has not seen, models are forced to generalize responses without being able to verify where it is accurate, relevant, or appropriate, which also contributes to hallucination. This issue also applies to other forms of problematic data, such as stereotypical data and biased data - due to the scale of the data used in the training of foundational models, it is impossible to fully or even significantly filter through and clean the training data. As a result, these issues rooted in problematic training data is likely to be present in many foundational models, which most fine-tuned model are based on and inherit from.

Another key reason source of hallucination is the prescence of duplicates in the training data ~\cite{lee2022deduplicating}, for which it is common for many training corpora to contain near-duplicate examples and long repetitive substrings. Not only does this lead to wasted compute resources in training, but it is observed to the model memorizing the common snippets. This can contribute to hallucination as the increased propensity to output memorized snippets may lead to the model generating them even in situations where it is not appropriate or factual to do so.

Furthermore, Ji et al.~\cite{Ji_2023} note that large language models exhibit Innate Divergence. This is because by nature, some natural language generation (NLG) tasks do not always have factual knowledge alignment between the source input text and the target reference, especially in situations that value diversity in the generated output. For instance, it may be acceptable for open-domain dialogue systems to respond in a subjective style,  as this can make generated dialogues more engaging and diverse. However, it has been observed that such dataset characteristics can lead to hallucinations. Such divergence can also lead to issues in the other direction, that is, the training data may contain pieces of information that are relatively informal and subjective, which a model may use in data generation even when prompted with a question that requires a factual answer.

With respect to models, Ji et al.~\cite{Ji_2023} point out that the training and inference of models are known to contribute to hallucination through a variety of factors, including Imperfect Representation Learning, Erroneous Decoding, Exposure Bias and Parametric Knowledge Bias. 

Imperfect Representation Learning occurs when an encoder is unable to map inputs to the appropriate internal embeddings, which could influence the degree of hallucination. This is because the encoder has the key role of encoding input text into meaningful representations, and when encoders learn wrong correlations between different parts of the training data, it could produce an erroneous generation that diverges from the input, which contributes to hallucination. 

Erroneous Decoding occurs when decoders focus on the wrong part of the encoded input source, leading to erroneous generation. It occurs also when decoders are designed in a way that improves the generation diversity, such as with changing top-p sampling, which is observed to be positively correlated with increased hallucination.

Exposure Bias occurs because of the discrepancy in decoding between training and inference time, which can also contribute to hallucination. This is rooted in the common practice of training decoders with teacher-forced MLE training, where the decoder is encouraged to predict the next token conditioned on the ground-truth prefix sequences. However, during the inference generation, the model generates the next token conditioned on the historical sequences previously generated by itself, which can contribute to hallucination through erroneous generation.

Parametric Knowledge Bias occurs because re-training of models on a large amount of data is known to result in the model memorizing knowledge in its parameters. While this is found to improve the performance in downstream tasks, it is also observed to contribute to hallucination, as models may prioritize parametric knowledge over the provided input. This also applies in generative question and answering, because if the phrasing of a question is too different from that of how knowledge was stored in the model, the intended answer may not be generated. Moreover, if the phrasing of a question is closer to that of another question that a model is more confident about, it may be the case that the model provides the answer for the other question and not the intended one, which can cause increased incorrect responses.


\section{Knowledge Mismatch Hypothesis}

The \textbf{hypothesis} examined in this paper is that \emph{there could be a mismatch between the knowledge that is fed to the model to fine-tune it and the knowledge that is already present in the graph, which could contribute to hallucination}.

At a small scale, through fine-tuning, the model learns a simple function that extracts knowledge from the model and outputs token predictions. For instance, fine-tuning a model with multiple prompts similar to the following prompt: "Q: Who is the leader of the United States? A: Joe Biden", which has already been learned by the model during training, would eventually lead the model to learn a function that predicts the right answer.

However, if we attempt to fine-tune the model on correct answers that have not been learned by the model, for instance, if the person creating fine-tuning instruction data knows about where Joe Biden was formerly a senator but the model does not, a kind of mismatch occurs. In this case, instead of training the model to output correct answers, this mismatch trains the model to guess answers for questions similar to the one that was newly introduced, because even if the question is not previously known, the fine-tuning teaches it to reply with a particular answer to such a question (instead of "I don't know"). In training the model to provide a response for something it did not learn before, it is hypothesized that the model is taught that a correct answer could be a piece of information that it did not learn before, and thus, it is taught to guess.

Furthermore, if the person creating fine-tuning instruction data does not know the president who succeeded Joe Biden, they will write the following prompt: "Q: Who succeeded Joe Biden? A: I don't know". In this case, they are fine-tuning the model to answer "I don't know" whereas the information is already known by the model. This leads to a mismatch. In this case, this mismatch trains the model to withhold information.

It is hypothesized that a mismatch occurs when there is a difference between the knowledge fed to the model to fine-tune it and the knowledge that is already present in the model. This mismatch either teaches the model to hallucinate or to withhold information.

\section{Testing the Hypothesis}
To test the hypothesis, we take two language models of different sizes, a small model and a large model, we fine-tune the small model with data generated using the large model (we will explain later how such data is generated). Since the large model has more knowledge compared to the small model, the data generated by the larger model is more likely to mismatch with the knowledge of the small model. We then measure the effect of this mismatch on the amount of hallucination in the small model. If the hypothesis is correct, then the small model will hallucinate more, since it was trained with data that has a mismatch with its knowledge.

More precisely, we take two models of different sizes, a small model and a large model, and fine-tune each model to abstain when it doesn't know the answer (i.e., answer "I don't know"). This is done by taking a dataset of questions and generating responses to its questions via prompted completions to create a fine-tuning dataset for the small and large models. \\

For example, for the question "Where was horse racing's Breeders' Cup held in 1988?":
\begin{enumerate}
  \item The model is first made to complete the prompt "Question: Where was horse racing's Breeders' Cup held in 1988? Answer:"
  \item Then, the generated response is compared with the list of correct answers provided by the dataset (a list of correct answers might be Churchill Downs, Louisville, Kentucky).
  \item If the generated response contains at least one correct answer, for instance, Churchill Downs, then it is marked as correct and the correct answer saved for the fine-tune response of the question. If it does not contain any correct answer, for instance, California, then it is marked as wrong and "I don't know" is saved as the fine-tune response of the question.
  \item The fine-tuning dataset is then created by using the questions and fine-tune response, which contains either correct answers or "I don't know". In doing this, the model is trained to abstain when it does not know the answer. 
\end{enumerate}

We then take the fine-tuning dataset created by the small model and the large model and use it to fine-tune the small model. Our hypothesis would suggest that the small model fine-tuned on the fine-tuning dataset generated by the large model will produce wrong answers more often.

\subsection{Experimental Setup}
\paragraph{Models}
We use LLaMA models by \cite{touvron2023llama} to test the hypothesis. We use the LLaMA 7B model as our small model and the LLaMA 13B model as our large model.

\paragraph{Data}
We use TriviaQA: A Large Scale Dataset for Reading Comprehension and Question Answering \cite{joshi-etal-2017-triviaqa} as our source of question and answer data. We split the data 80/20 between the training set and the testing set.

\paragraph{Fine-tuning}
We use parameter-efficient fine-tuning ~\cite{peft} to tune the model using fine-tuning code from Project Baize \cite{xu2023baize}.

\paragraph{Evaluation}
We evaluate the model by making the fine-tuned small models generate responses to the unseen test split of the data.

\subsection{Results}
After evaluating the fine-tuned models on the unseen test set, we found that the small model fine-tuned on the data generated by the large model produces more wrong answers than the small model fine-tuned on data generated by the small model - an average of 125\% and median of 107\% increase in wrong answers.\\

The difference between the two models can be seen in Table~\ref{table1}. For example in the 5th row of the table, the small model (7B) fine-tuned on data generated by the small model, generates 426 wrong answers. The same model, when fine-tuned on data generated by the large model (13B), generates 1134 wrong answers (as shown in the 10th row). This confirms the hypothesis. \\

Table~\ref{table2} shows the increase in the number of wrong responses between the two models. Row 5 of the table, for example, shows that there is an increase of $166.2\%$ in wrong response in the "7B model fine-tuned on 13B" compared to the "7B model fine-tuned on 7B".

\begin{table*}[ht!]
  \caption{Evaluation of Fine-tuned Models on Unseen Test Set}
  \label{sample-table1}
  \centering
  \begin{tabular}{lrrrr}
    \toprule
Model and Finetune  & \multicolumn{1}{l}{Epochs} & \multicolumn{1}{l}{Wrong} & \multicolumn{1}{l}{Correct } & \multicolumn{1}{l}{IDK} \\
    \midrule
7B Finetuned on 7B  & 1                                                  & 675                               & 7205                                & 9644                                  \\
7B Fine-tuned on 7B  & 2                                                  & 1072                              & 7019                                & 9433                                  \\
7B Fine-tuned on 7B  & 3                                                  & 704                               & 7362                                & 9458                                  \\
7B Fine-tuned on 7B  & 4                                                  & 643                               & 7016                                & 9865                                  \\
7B Fine-tuned on 7B  & 5                                                  & 426                               & 6614                                & 10484                                 \\
7B Fine-tuned on 13B & 1                                                  & 1168                              & 8390                                & 7966                                  \\
7B Fine-tuned on 13B & 2                                                  & 1546                              & 7904                                & 8074                                  \\
7B Fine-tuned on 13B & 3                                                  & 1459                              & 7495                                & 8570                                  \\
7B Fine-tuned on 13B & 4                                                  & 2150                              & 7887                                & 7487                                  \\
7B Fine-tuned on 13B & 5                                                  & 1134                              & 7475                                & 8915                                 \\
    \bottomrule
  \end{tabular}
  \label{table1}
\end{table*}

\begin{table*}[ht!]
  \caption{Change in responses between fine-tunes of 7B and 13B data }
  \label{sample-table2}
  \centering
  \begin{tabular}{lrrrr}
    \toprule
Epochs                & \multicolumn{1}{l}{\begin{tabular}[c]{@{}l@{}}\% increase in\\ wrong responses\end{tabular}} & \multicolumn{1}{l}{\begin{tabular}[c]{@{}l@{}}\% increase in correct\\ responses\end{tabular}} & \multicolumn{1}{l}{\begin{tabular}[c]{@{}l@{}}\% decrease in\\ IDK responses\end{tabular}} & \multicolumn{1}{l}{} \\

    \midrule
\multicolumn{1}{r}{1} & 73.04                                                                                       & 16.45                                                                                          & 17.4                                                                                       &                      \\
\multicolumn{1}{r}{2} & 44.22                                                                                       & 12.61                                                                                          & 14.41                                                                                      &                      \\
\multicolumn{1}{r}{3} & 107.24                                                                                      & 1.81                                                                                           & 9.39                                                                                       &                      \\
\multicolumn{1}{r}{4} & 234.37                                                                                      & 12.41                                                                                          & 24.11                                                                                      &                      \\
\multicolumn{1}{r}{5} & 166.2                                                                                       & 13.02                                                                                          & 14.97                                                                                      &                      \\
        Average & 125.01 & 11.26 & 16.05 \\ 
        Median & 107.24 & 12.61 & 14.97 \\ 
        \hline 
    \bottomrule
  \end{tabular}
  \label{table2}
\end{table*}

\begin{figure*}
    \centering
    \caption{Comparison of Wrong Answers}
    \label{Comparison of Wrong Answers}
\begin{tikzpicture}
\begin{axis}[
    ybar=3pt,  
    enlargelimits=0.15,
    legend style={at={(0.5,-0.2)},
    anchor=north,legend columns=-1},
    ylabel={\# of Wrong answers},
    symbolic x coords={1,2,3,4,5},
    xtick=data,
    nodes near coords,
    nodes near coords align={vertical},
    x tick label style={rotate=45,anchor=east},
    bar width = 3pt,  
    width=8cm,  
    height=8cm,  
    xlabel={Epochs},
]
\addplot coordinates {(1,675) (2,1072) (3,704) (4,643) (5,426)};
\addplot coordinates {(1,1168) (2,1546) (3,1459) (4,2150) (5,1134)};

\legend{7B Fine-tuned on 7B, 7B Fine-tuned on 13B}
\end{axis}
\end{tikzpicture}
\end{figure*}
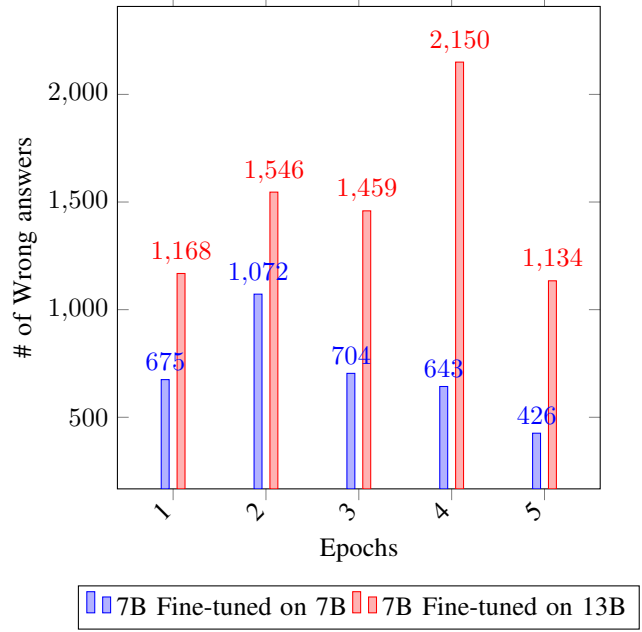

A comparison of the number of wrong answers in the two models is summarized in Figure \ref{Comparison of Wrong Answers}.

\section{The Effects of Fine-tuning on Data Obtained from a Larger Model}
Our results show that a small model fine-tuned on data generated by a larger model provides significantly more wrong answers compared to small models fine-tuned on data generated by the small model. This suggests that fine-tuning on data generated by large models may have the unintended consequence of increasing the hallucination through an increase in factually incorrect responses in questioning and answering settings. \\

We also observe that there are fewer "I don't know" responses in the responses of the model fine-tuned on data generated by a larger model compared to that of the model fine-tuned on data generated by the smaller model. We believe that this is most likely due to the fact that the fine-tuning dataset created by the base 13B model has fewer "I don't know" responses compared to the one generated by the base 7B model because the base 13B model performs better than the base 7B model and thus had fewer wrong answers that were mapped to "I don't know". \\

In addition, we find that there are more correct answers in the model fine-tuned with the data generated by that larger model than that of the small model. This is in line with the evaluations of many of the small-scale larger model data fine-tune models released recently. This is likely due to the fact that despite causing some knowledge mismatch, fine-tuning still contributes to improved model performance.

\section{Limitations}
Knowledge mismatch through model fine-tuning is only one of many potential reasons for large language model hallucination. As such, there may be other factors that may also contribute to model hallucination that our analysis does not discuss. \\

Given the resources available, we only tested the hypothesis on the 7B and 13B variants of the LLaMA model. While we believe that our hypothesis should likely still hold for other model combinations, we were not able to test this because of resource constraints. \\

It is also worth noting that while the TriviaQA dataset is a relatively large and widely used question and answering dataset, is not fully processed and cleaned. In particular, a few answers in the dataset are not correct, and some answers do not list all possible answers and answer formats. As such this may contribute to a few incorrect evaluations where the model was correct but the expected answer was incorrect. 

\section{Related Work}
Most model evaluations, including many fine-tuned models trained using data from larger models generally report their hallucination as a key limitation and highlight the importance of using quality data in model tuning to achieve better performance. \\

Most top AI research organizations also generally report model hallucination and the techniques that they use to lower and limit hallucination. Currently, there exist many attempts to create, leverage and analyze strategies to reduce hallucination, such as RLHF and RLAIF. \\

For instance, ~\cite{ouyang2022training} observe that making language models bigger does not inherently make them better at following a user’s intent, as large language models can still generate outputs that are untruthful, toxic, or simply not helpful to the user. They note that large language models are not necessarily aligned with their users and propose RLHF, also known as Reinforcement Learning with Human Feedback, which finetunes models with human feedback to address the issue. They observe that in human evaluations outputs from a smaller model (InstructGPT 1.3B) that leverages RLHF are preferred to outputs from a model 100x the size (GPT3). They also find that InstructGPT models show improvements in truthfulness and reductions in toxic output generation while having minimal performance regressions on public NLP datasets, which suggests that RLHF can contribute to reducing hallucination. ~\cite{bai2022training} and ~\cite{thoppilan2022lamda} apply a similar strategy and also find less hallucination and more helpful, safe, and aligned models. \\

Another common technique to lessen hallucination and create more aligned models is RLAIF or Reinforcement Learning with Artificial Intelligence Feedback. ~\cite{bai2022constitutional} implement this strategy through Constitutional AI, for which the only human oversight provided is a list of rules or principles, and the AI model is used to train itself through supervised training and reinforcement learning (RL). In the supervised phase, self-critiques and revisions are generated from the model, which are then revised and used to fine-tune the original model. In the RL phase, samples are generated from the finetuned model, which is fed to a model to evaluate which of the two samples is better, and then used to train a preference model from this dataset of AI preferences. They then engage in RLAIF by training with RL using the preference model as the reward signal. They find that doing this reduces harmfulness similar to human feedback. \\

There has also been some work in studying the effects of novel training and training data generation through techniques such as Contrastive Learning and Evol-Instruct. \\

~\cite{sun2022contrastive} study the use of contrastive learning as a potential way to address hallucination. In particular, they propose Mixed Contrastive Learning (MixCL) to alleviate the hallucinations of LM-based dialogue agents. This technique explicitly samples the most confusing knowledge to the model and reduces its generation probability by contrasting it with the ground truth. They find that this technique significantly improves dialogue performance and lessens the hallucination of LMs, based on ablation studies and human evaluation. \\

~\cite{xu2023wizardlm} focus on creating better data for model fine-tuning. In particular, they propose Evol-Instruct as a technique to generate high-quality data for fine-tuning that can lead to more helpful models. They note that manually creating instruction data is not only time-consuming and labor-intensive but also difficult, as humans may struggle to produce high-complexity instructions. In their study, they propose a method to create instruction data with varying levels of complexity using LLMs instead of humans. Using an initial set of instructions, they use Evol-Instruct to rewrite them step by step into more complex instructions, then mix the generated instruction data to fine-tune a foundational model. They find that the resulting fine-tuned model is superior to top fine-tuned models of a similar size, including the Vicuna, which is fine-tuned on ShareGPT data. They also find that the fine-tuned model is even preferred over ChatGPT (a significantly larger proprietary model) in high-complexity tasks. This technique of Evol-Instruct can be said to be in line with the proposed hypothesis as it leverages the outputs generated by the same model for fine-tuning, which means that there should be significantly less knowledge mismatch in the fine-tuning, which may have contributed it its superior performance.

\section{Conclusions}
We have explored the hypothesis that a mismatch between the knowledge that is fed to the model to fine-tune it and existing knowledge in a graph can contribute to the propensity of a model to hallucinate. Through experiments on fine-tuning models on data generated from other models, we find that models fine-tuned on data generated by larger models produce more wrong answers on questions that it has not seen before. Studying potential causes of hallucination will hopefully inspire further work on better techniques to create more helpful, truthful, and harmless models.

\section{Acknowledgments}
This research has been partly supported by the Center for Artificial Intelligence and Robotics (CAIR) and Center for Cyber Security (CCS) at New York University Abu Dhabi. The research was carried out on the High-Performance Computing resources at New York University Abu Dhabi.


\begin{thebibliography}{00}
\bibitem{touvron2023llama} H. Touvron, T. Lavril, G. Izacard, X. Martinet, M.-A. Lachaux, T. Lacroix, B. Rozière, N. Goyal, E. Hambro, F. Azhar, A. Rodriguez, A. Joulin, E. Grave, and G. Lample, ``LLaMA: Open and Efficient Foundation Language Models,'' arXiv:2302.13971, 2023.
\bibitem{hu2021lora} E. J. Hu, Y. Shen, P. Wallis, Z. Allen-Zhu, Y. Li, S. Wang, L. Wang, and W. Chen, ``LoRA: Low-Rank Adaptation of Large Language Models,'' arXiv:2106.09685, 2021.
\bibitem{xu2023baize} C. Xu, D. Guo, N. Duan, and J. McAuley, ``Baize: An Open-Source Chat Model with Parameter-Efficient Tuning on Self-Chat Data,'' arXiv:2304.01196, 2023.
\bibitem{sun2022contrastive} W. Sun, Z. Shi, S. Gao, P. Ren, M. de Rijke, and Z. Ren, ``Contrastive Learning Reduces Hallucination in Conversations,'' arXiv:2212.10400, 2022.
\bibitem{bang2023multitask} Y. Bang, S. Cahyawijaya, N. Lee, W. Dai, D. Su, B. Wilie, H. Lovenia, Z. Ji, T. Yu, W. Chung, Q. V. Do, Y. Xu, and P. Fung, ``A Multitask, Multilingual, Multimodal Evaluation of ChatGPT on Reasoning, Hallucination, and Interactivity,'' arXiv:2302.04023, 2023.
\bibitem{joshi-etal-2017-triviaqa} M. Joshi, E. Choi, D. Weld, and L. Zettlemoyer, ``TriviaQA: A Large Scale Distantly Supervised Challenge Dataset for Reading Comprehension,'' in Proc. 55th Annu. Meeting of the Assoc. for Comput. Linguistics, pp. 1601--1611, July 2017.
\bibitem{dolly} M. Conover, M. Hayes, A. Mathur, X. Meng, J. Xie, J. Wan, A. Ghodsi, P. Wendell, and M. Zaharia, ``Hello Dolly: Democratizing the Magic of ChatGPT with Open Models,'' Databricks Blog, 2023. [Online]. Available: \url{https://www.databricks.com/blog/2023/03/24/hello-dolly-democratizing-magic-chatgpt-open-models.html}
\bibitem{alpaca-lora} E. Wang, ``Alpaca-LoRA,'' GitHub repository, 2023. [Online]. Available: \url{https://github.com/tloen/alpaca-lora}
\bibitem{gpt4all} Y. Anand, Z. Nussbaum, B. Duderstadt, B. Schmidt, and A. Mulyar, ``GPT4All: Training an Assistant-Style Chatbot with Large Scale Data Distillation from GPT-3.5-Turbo,'' GitHub repository, 2023. [Online]. Available: \url{https://github.com/nomic-ai/gpt4all}
\bibitem{alpaca} R. Taori, I. Gulrajani, T. Zhang, Y. Dubois, X. Li, C. Guestrin, P. Liang, and T. B. Hashimoto, ``Stanford Alpaca: An Instruction-Following LLaMA Model,'' GitHub repository, 2023. [Online]. Available: \url{https://github.com/tatsu-lab/stanford_alpaca}
\bibitem{vicuna2023} W.-L. Chiang, Z. Li, Z. Lin, Y. Sheng, Z. Wu, H. Zhang, L. Zheng, S. Zhuang, Y. Zhuang, J. E. Gonzalez, I. Stoica, and E. P. Xing, ``Vicuna: An Open-Source Chatbot Impressing GPT-4 with 90\% ChatGPT Quality,'' 2023. [Online]. Available: \url{https://vicuna.lmsys.org/}
\bibitem{koala_blogpost_2023} X. Geng, A. Gudibande, H. Liu, E. Wallace, P. Abbeel, S. Levine, and D. Song, ``Koala: A Dialogue Model for Academic Research,'' Blog post, Apr. 2023. [Online]. Available: \url{https://bair.berkeley.edu/blog/2023/04/03/koala/}
\bibitem{bai2022constitutional} Y. Bai, S. Kadavath, S. Kundu, A. Askell, J. Kernion, A. Jones, A. Chen, A. Goldie, A. Mirhoseini, C. McKinnon, C. Chen, C. Olsson, C. Olah, D. Drain, D. Hernandez, D. Li, E. Tran-Johnson, E. Perez, J. Kerr, J. Mueller, J. Ladish, K. Ndousse, K. Lukosuite, L. Lovitt, M. Sellitto, N. Elhage, N. Schiefer, N. Mercado, N. DasSarma, R. Larson, S. Ringer, S. Johnston, S. Kravec, S. Fort, T. Lanham, T. Telleen-Lawton, T. Conerly, T. Henighan, T. Hume, S. R. Bowman, Z. Hatfield-Dodds, B. Mann, D. Amodei, N. Joseph, and S. McCandlish, ``Constitutional AI: Harmlessness from AI Feedback,'' arXiv:2212.08073, 2022.
\bibitem{ouyang2022training} L. Ouyang, J. Wu, X. Jiang, D. Almeida, C. L. Wainwright, P. Mishkin, C. Zhang, S. Agarwal, K. Slama, A. Ray, J. Schulman, J. Hilton, F. Kelton, L. Miller, M. Simens, A. Askell, P. Welinder, P. Christiano, J. Leike, and R. Lowe, ``Training Language Models to Follow Instructions with Human Feedback,'' arXiv:2203.02155, 2022.
\bibitem{bai2022training} Y. Bai, A. Jones, K. Ndousse, A. Askell, A. Chen, N. DasSarma, D. Drain, S. Fort, N. Joseph, S. Kadavath, J. Kernion, T. Henighan, T. Hume, Z. Hatfield-Dodds, D. Hernandez, S. Johnston, S. Kravec, L. Lovitt, N. Nanda, C. Olsson, D. Amodei, T. Brown, J. Clark, S. McCandlish, C. Olah, B. Mann, and J. Kaplan, ``Training a Helpful and Harmless Assistant with Reinforcement Learning from Human Feedback,'' arXiv:2204.05862, 2022.
\bibitem{dziri2022origin} N. Dziri, S. Milton, M. Yu, O. Zaiane, and S. Reddy, ``On the Origin of Hallucinations in Conversational Models: Is it the Datasets or the Models?,'' arXiv:2204.07931, 2022.
\bibitem{peft} S. Mangrulkar, S. Gugger, L. Debut, Y. Belkada, and S. Paul, ``PEFT: State-of-the-art Parameter-Efficient Fine-Tuning Methods,'' GitHub repository, 2022. [Online]. Available: \url{https://github.com/huggingface/peft}
\bibitem{Ji_2023} Z. Ji, N. Lee, R. Frieske, T. Yu, D. Su, Y. Xu, E. Ishii, Y. J. Bang, A. Madotto, and P. Fung, ``Survey of Hallucination in Natural Language Generation,'' ACM Comput. Surveys, vol. 55, no. 12, pp. 1--38, Mar. 2023.
\bibitem{dziri2022evaluating} N. Dziri, H. Rashkin, T. Linzen, and D. Reitter, ``Evaluating Attribution in Dialogue Systems: The BEGIN Benchmark,'' arXiv:2105.00071, 2022.
\bibitem{thoppilan2022lamda} R. Thoppilan, D. De Freitas, J. Hall, N. Shazeer, A. Kulshreshtha, H.-T. Cheng, A. Jin, T. Bos, L. Baker, Y. Du, Y. Li, H. Lee, H. S. Zheng, A. Ghafouri, M. Menegali, Y. Huang, M. Krikun, D. Lepikhin, J. Qin, D. Chen, Y. Xu, Z. Chen, A. Roberts, M. Bosma, V. Zhao, Y. Zhou, C.-C. Chang, I. Krivokon, W. Rusch, M. Pickett, P. Srinivasan, L. Man, K. Meier-Hellstern, M. Ringel Morris, T. Doshi, R. Delos Santos, T. Duke, J. Soraker, B. Zevenbergen, V. Prabhakaran, M. Diaz, B. Hutchinson, K. Olson, A. Molina, E. Hoffman-John, J. Lee, L. Aroyo, R. Rajakumar, A. Butryna, M. Lamm, V. Kuzmina, J. Fenton, A. Cohen, R. Bernstein, R. Kurzweil, B. Aguera-Arcas, C. Cui, M. Croak, E. Chi, Q. Le, "LaMDA: Language Models for Dialog Applications," arXiv preprint arXiv:2201.08239, 2022.
\bibitem{xu2023wizardlm} C. Xu, Q. Sun, K. Zheng, X. Geng, P. Zhao, J. Feng, C. Tao, D. Jiang, "WizardLM: Empowering Large Language Models to Follow Complex Instructions," arXiv preprint arXiv:2304.12244, 2023.
\bibitem{lee2022deduplicating} K. Lee, D. Ippolito, A. Nystrom, C. Zhang, D. Eck, C. Callison-Burch, N. Carlini, "Deduplicating Training Data Makes Language Models Better," arXiv preprint arXiv:2107.06499, 2022.


\end{thebibliography}
\end{document}